\documentclass[nonacm]{acmart}
\AtBeginDocument{%
  \providecommand\BibTeX{{%
    \normalfont B\kern-0.5em{\scshape i\kern-0.25em b}\kern-0.8em\TeX}}}

\usepackage{xspace}
\newcommand{\creators}{\textsc{Creators}\xspace}
\newcommand{\creator}{\textsc{Creator}\xspace}
\newcommand{\referrents}{\textsc{Referrents}\xspace}
\newcommand{\referrent}{\textsc{Referrent}\xspace}

\newcommand{\addressees}{\textsc{Addressees}\xspace}

\newcommand{\audience}{\textsc{Audience}\xspace}
\newcommand{\moderators}{\textsc{Moderators}\xspace}
\usepackage{multirow}

\usepackage[multiple]{footmisc}

\begin{document}

  
\title{A Human Rights-Based Approach to Responsible AI}



\author{Vinodkumar Prabhakaran}
\affiliation{%
  \institution{Google Research}
   \country{USA}
}
\email{vinodkpg@google.com}

\author{Margaret Mitchell}
\affiliation{%
  \institution{Hugging Face}
   \country{USA}
  }
\email{margarmitchell@gmail.com}

\author{Timnit Gebru}
\affiliation{%
  \institution{DAIR}
   \country{USA}
  }
\email{tgebru@gmail.com}

\author{Iason Gabriel}
\affiliation{%
  \institution{Deepmind}
   \country{UK}
  }
\email{iason@deepmind.com}
\begin{abstract}

Research on fairness, accountability, transparency and ethics of AI-based interventions in society has gained much-needed momentum in recent years. However it lacks an explicit alignment with a set of normative values and principles that guide this research and interventions. Rather, an implicit consensus is often assumed to hold for the values we impart into our models – something that is at odds with the pluralistic world we live in. In this paper, we put forth the doctrine of universal human rights as a set of globally salient and cross-culturally recognized set of values that can serve as a grounding framework for explicit value alignment in responsible AI – and discuss its efficacy as a framework for civil society partnership and participation. We argue that a human rights framework orients the research in this space away from the machines and the risks of their biases, and towards humans and the risks to their rights, essentially helping to center the conversation around who is harmed, what harms they face, and how those harms may be mitigated.

\end{abstract}

\maketitle

\newif\ifcasestudyinmaintext

\casestudyinmaintextfalse

\section{Introduction}
\label{sec_intro}

Fairness, accountability, transparency and ethics (FATE) research in AI contends with questions around how AI models might be problematically biased, unfair, or unethical, and how to make them ``fairer'' and more ``ethical''. However, the research community working in this area greatly lacks in terms of geo-cultural diversity \cite{freire2020measuring}, resulting in the research being primarily framed in the Western context,\footnote{In this paper, we use \textit{Western} or \textit{the West} to refer to the regions, nations and states, consisting of Europe, the United States and Canada, and Australasia and their shared norms, values, customs, religious beliefs, and political systems \cite{kurth2003western}.} by researchers mostly situated in Western institutions/organizations, to mitigate social injustices prevalent in the West, using data from the West, and implicitly imparting Western value systems \cite{sambasivan2021india}. On the other hand, these research insights are meant to intervene on platforms that are globally present, serving a global population from diverse societies, cultures and values, with their own forms of injustices.

A core concern in this arrangement is that of value imposition, where local values, i.e., values that are local to the regions where the interventions are built, implicitly shape and inform global systems without any or much room for discussion or contestation from those affected by those interventions. More specifically, interventions designed to address FATE failures necessarily impart a normative value system, but the values that guide the proposed solutions are rarely recognized as sites of contestation. This is problematic because while there may be ethical principles for ML that garner a degree of consensus across different value systems, in a pluralistic world this consensus is not something that should be assumed. Instead, we need to be explicit about the values that underpin the quest for ethical and just AI, and to cultivate an active debate about those values, critically examining and evaluating claims about them \cite{gabriel2020artificial}.

Another shortcoming of not being explicit about what normative value systems shape the interventions is the vagueness it entails, 
making it harder to arrive at a common vocabulary and shared understanding between computer scientists and civil society. Such a shared understanding is crucial to bridge the gap between research and practice, especially in a way that effectively supports the priorities of the latter constituency. 
This is especially important given the need for critical examinations that require deeper understanding of the societal contexts in which interventions are envisioned \cite{benthall2019racial,hanna2020towards,birhane2020towards,sambasivan2021india} and the need for participatory methods to incorporate marginalized stakeholder perspectives \cite{martin2020participatory,katell2020toward,donia2021co} when shaping these interventions.

In this paper, we argue that the doctrine of \textit{human rights} can serve as a starting point to help address some of these gaps. The role of human rights as a legal framework, through which AI-related harms may be identified, has been explicitly evoked by (or implicitly shaped) various AI accountability initiatives within the industry as well as governmental and civil society bodies (e.g., \cite{raso2018artificial,marda2018artificial,cath2018governing,mcgregor2019international}). However, we take a broader view of human rights and argue that, in addition to this legal role, they may play three further functions. First, human rights, understood as a set of moral claims, have a measure of intercultural and cross-cultural validity, which means that they can support value alignment for AI systems across a range of different national and social contexts. Second, they can help illuminate the concurrent responsibilities of various actors, given that they apply to states, organizations and individuals. Third, they provide a shared vocabulary and framework that technologists and practitioners may productively invoke to address the claims and concerns of the global civil society and the people impacted by these technologies. To clarify this distinction, we start by distinguishing three different aspects of human rights --- (1) as a set of moral claims, (2) as a legal regime and set of instruments, and (3) as a cultural practice and global social movement. 
We then delve deeper into three specific rights enshrined in the Universal Human Rights Declaration (UDHR),\footnote{\url{https://www.un.org/en/about-us/universal-declaration-of-human-rights}} in order to illustrate how human rights might shape responsible AI development and deployments. 
\ifcasestudyinmaintext
Finally, using a set of specific hypothetical scenarios within the AI application context of content moderation, we demonstrate how a human rights approach can shape decisions on application development, and explore how different choices may protect some rights while risking others. 
\fi
While we acknowledge that a human rights perspective is not a panacea for addressing all issues in AI use, and that the provenance of UDHR in particular has been contested, we argue that a cross-cultural set of human rights, exemplified here by UDHR, can be a grounding framework for value alignment in AI.

\section{Human Rights and Ethical AI}

From a philosophical perspective, human rights are the fundamental rights every human being holds simply by virtue of their birth, regardless of their age, ethnic origin, location, language, religion, nationality, ethnicity, or any other status.\footnote{\url{https://www.ohchr.org/en/issues/pages/whatarehumanrights.aspx}} The doctrine of human rights---which maps out their content, meaning, and consequences---has a number of different parts. 

First and foremost, human rights can be understood as a set of \textit{moral claims} anchored in the notion that human life has value and that there are aspects of personhood that must be protected. These claims can be grounded in a number of different ways but they tend to draw support from recognition that there are a set of core human interests that are widely shared, from common appreciation of the need to respect human autonomy and freedom, and from shared recognition of the dignity of human life \cite{griffin2009human,beitz2009idea,donnelly2013universal,shue2020basic,freeman2022human}. In each case, there is a consideration that is strong enough to create a duty on other actors to only treat people in certain ways. In the words of Henry Shue, those who possess human rights can make justified demands that the actual enjoyment of a good be socially guaranteed against standard threats to those things \cite{shue2020basic}. Taken together, the idea of human rights holds that we owe it to each other to build a world in which the ability to enjoy certain goods, such as physical security, health and education, are widely enjoyed by all. 

Second, human rights are part of a \textit{legal regime and set of instruments} that aim (in part) to make these values a reality \cite{buchanan2013heart}.\footnote{Legal regimes refers to the different bodies of canonical law that apply in a context; for instance, there is an international legal regime, an EU legal regime, a US legal regime etc., which may overlap but are not necessarily precisely compatible.} While the idea of universal rights has a long intellectual lineage, the modern conception of human rights emerged largely after the Second World War in response to the atrocities that had been committed and the moral trauma of the Holocaust \cite{claude1992human}. This experience helped seed the desire to forge international agreement around the protection states owed to their citizens (and more widely), culminating in the adoption of the Universal Declaration of Human Rights (UDHR) by the United Nations General Assembly in 1948. The UDHR consists of 30 different articles affirming an individual's rights, such as the right to life, liberty and security, right to privacy, right to be free of discrimination, and right to freedom of expression --- rights that are critically relevant to building AI-based interventions responsibly. Drafted by a committee that included representatives from China, Chile, the Soviet Union, Lebanon, and India (in addition to the Western powers at the time), the UDHR was subsequently augmented by the International Covenant on Civil and Political Rights (ICCPR) and the International Covenant on Economic, Social and Cultural Rights (ICESCR) in 1966. These agreements incorporated insights from a still larger group of countries after the widespread success of decolonization movements. They now form part of customary international law and have been adopted by a number of national and transnational legal bodies such as the International Criminal Court (ICC).

Third, human rights are part of a \textit{cultural practice and global social movement} that focuses on advocacy, empowerment and critique of existing institutions. From the very beginning, civil society has played a vital role in developing human rights doctrine, embedding norms, and monitoring outcomes \cite{keck1998activists}. As human rights advocates have long understood, it is not enough for rights to exist only in moral or juridical form: people also need to be informed about their rights and exercise them, if they are to have a significant bearing upon outcomes. Human rights advocacy has had a number of important successes, including in the field of women’s rights, indigenous rights, and disability rights, helping to shape global norms around how people may and may not be treated \cite{risse1999power}. Civil society organizations have also explicitly and successfully used the human rights mechanisms to contest government policies (e.g., by making submissions to the UN's Universal Periodic Review \cite{gomez2018singapore}).

Efforts are now underway to incorporate human rights into the design of AI systems, and to identify novel ways in which AI can support human rights practices. In applied contexts, there have been a number of attempts to integrate AI with human rights monitoring. These include efforts to improve human rights reporting in conflict zones using automated analysis of satellite imagery \cite{livingston2019future}. Additionally, in the context of discussions about fairness and AI ethics, human rights frameworks have been proposed as a tool or mechanism to promote greater accountability to those in need  \cite{raso2018artificial,marda2018artificial,cath2018governing,mcgregor2019international}. In particular human rights advocates have criticised the tendency, sometimes present in FATE discourse, to imply that the relevant ethical norms for AI technology need to be discovered for the very first time --- or to focus on sophisticated statistical analysis of algorithms without paying due attention to the way in which their deployment may result in actual societal harms \cite{selbst2019fairness}. Against these viewpoints, human rights doctrine contains a set of well-established principles that are robustly centered upon human vulnerabilities and human needs. Finally, there has been notable engagement with specific challenges to human rights posed by new technologies such as fake news moderation \cite{marda2018wisdom} and predictive policing \cite{marda2020data}. For example, Data \& Society organized a multidisciplinary workshop in April 2018 exploring how the human rights framework can effectively inform, shape, and govern AI research, development, and deployment. More recently, \cite{cath2020leap} organized tutorials at the FAT* 2020 conference on this topic.\footnote{The conference has since been renamed to FAccT.} We draw inspiration from these groundbreaking efforts, and argue that more substantial engagement from the machine learning research community with the human rights paradigm can address three major challenges that the field of ML faces: alignment, the allocation of responsibilities, and participation.

\subsection{As a Basis for Cross-Cultural Value Alignment}
\label{hr_value}

Any effort to try and create ethical AI involves imparting a set of human values on the behavior of the AI-based systems under consideration, many of which are deployed across the world. While there has been plenty of research within AI sub-communities such as AI safety \cite{arnold2017value}, machine learning \cite{hadfield2016cooperative}, and natural language processing \cite{card2020consequentialism,prabhumoye2021case}, on the challenges involved in encoding human values into AI systems, less attention has been paid to the normative question of \textit{which} set of values/principles should be encoded. Sometimes this question is overlooked altogether; in other cases system designers proceed on the basis of a preferred principle or theory. Both approaches are problematic given the wide variation in the moral beliefs people actually hold.\footnote{For instance, \url{www.worldvaluessurvey.org/}} Indeed, since most FATE research in AI is situated in the West, it does not meaningfully engage with the conditions, values, and histories of non-West contexts \cite{jobin2019global,sambasivan2021india,png2022tensions}. As Birhane and Cummins note in \cite{relational_ethics}, ``it is possible that what is considered ethical currently and within certain domains for certain societies will not be received similarly at a different time, in another domain, or for a different society.'' Thus, the Western values implicitly encoded into AI systems may be at odds with other value systems, creating the risk of problematic value imposition when these technologies are deployed globally \cite{gabriel2020artificial}.


In order to address this challenge, Gabriel \cite{gabriel2020artificial} argues that we need to identify fair processes for selecting values to encode in AI systems. In this context, one promising approach focuses on the possibility of identifying an “overlapping consensus” between the different moral belief systems around the world. To make progress in this direction we need to ask: are there any ideals that command widespread, or even global, assent? Although no candidate doctrines is without limitations, with respect to cross-cultural validity, the doctrine of universal human rights is particularly promising \cite[p.~150]{donnelly2007relative}. Indeed, while the modern notion of human rights has the specific historical lineage that we have outlined, a strong case can be made for the notion that these rights are no longer time-bound or culturally parochial. To start with, there is evidence that these beliefs find a degree of cross-cultural support in African, Islamic, Western, and Confucian traditions of thought \cite[p335-343]{cohen2010arc}. Moreover, human rights have been adopted and \textit{actively claimed} by people around the world across a wide range of contexts ranging from indigenous rights movements \cite{pentassuglia2011towards} to the Arab Spring \cite{harrelson2014you}. Hence, we believe human rights can serve as a legitimate goal for building value-aligned AI systems, a requirement that entails both that they need to respect human rights directly and also that they are deployed in ways that strengthen human rights practice.


Yet, the proposed focus on human rights faces four objections. First, it might be thought to lead to certain knowledge gaps: can a human rights framework adequately factor in collective goods, for example, or concerns about distributive justice? This criticism has often been made by Marxist scholars and contemporary critics of human rights doctrine such as Samuel Moyn \cite{moyn2018not}. In certain respects, the concerns they raise are well-founded: we agree that there may well be goods and considerations that are not adequately expressed in the language of rights \cite{raz1986right}. Yet, our claim is not that human rights capture the full space of AI ethics but only that they capture a set of particularly important claims --- a ‘morality of the depths’ \cite{shue2020basic} --- for which there is widespread support. Understood in this way, respect for human rights would be necessary but not sufficient to ensure the ethical design and deployment of AI systems.  

Second, scholars have challenged the purported ``universalilty'' of human rights by mobilising arguments that draw upon cultural relativism \cite{wai1979human}. Proponents of this viewpoint note that moral norms vary significantly across cultures, and  that what is ``regarded as a human rights violation in one society may properly be considered lawful in another, and Western ideas of human rights should not be imposed upon Third World societies'' \cite{teson1985international}. However, as Donnelly notes, a more nuanced analysis of this matter resists the reduction of human rights doctrine to this single axis, and asks instead ``\textit{how} human rights are (and are not) universal and how they are (and are not) relative'' \cite{donnelly2013universal}. In this context, he notes that UDHR has a very strong claim to what he terms ``relative universality'', insofar as it represents a minimal response to basic cross-cultural human values and the threats to human dignity posed by modern institutions. At the same time, from a practical standpoint, the ways in which human rights interface with more specific cultural values is an important aspect to consider in the doctrine's integration into AI ethics \cite{wong2020cultural}. For instance, human rights doctrine may serve as an useful starting point that provides a baseline set of values that enjoy cross-cultural recognition, that can then be built on for specific contexts, especially when rights of specific communities are in question.


Third, despite what has been said, we might worry that human rights are still geographically parochial, or worse, a form of neo-imperialism \cite{wall1998human}. In this context, it is important to recognize the historicity of human rights, the evolving character of human rights discourse, and the political undercurrent that influence how these norms are operationalized in practice. There is certainly evidence that human rights claims have been deployed in a strategic or politicized way by Western states or NGOs at certain times, and that they still operate under the burden of this legacy today \cite{hardt2000empire}. However, this does not detract from the reality of their widespread support, or from the fact that they have been used equally in an opposing manner --- to oppose authoritarian regimes, and in anti-colonial movements to resist external intervention \cite{burke2011decolonization,jensen2016making}.
Moreover, Kathryn Sikkink notes that “voices and actors from the Global South were deeply involved in demanding the international protection of human rights and in building the institutions that started to make enforcement of these rights possible” \cite[p.~25]{Sikkink2017}. As a result of this process of contestation, something of value has emerged: modern human rights doctrine is often invoked by people to resist interference with their political rights or ability to govern themselves. 

Finally, it might be thought that a human rights framework is too anthropocentric, emphasising the value of  human life while neglecting non-human life and other bearers of value. However, as we noted before, insisting on the importance of human rights is not the same as asserting that only human rights matter: we agree other things matter as well. Moreover, we suggest that a renewed focus on human rights --- and on the ways in which people can be seriously harmed --- is particularly important for AI research at the present moment given the tendency to approach ethical challenges as technical problems \cite{birhane2020robot}. Afterall, a great deal has been written about the formal properties of models and the biases they embody, however, the need to better understand the harms these models cause is critical \cite{barocas2017problem,altman2018harm}. By way of illustration, Blodgett et al. conducted a survey of 146 research papers analyzing bias in natural language processing models and found only limited engagement with the question of why bias is harmful, in what ways, and to whom \cite{blodgett2020language}. As a consequence, it is hard to be confident that proposed mitigations will have a positive impact on those who encounter these harms. By way of contrast, a human rights based approach to fairness research can help forge a stronger connection between models, the socio-technical systems they operate in, and salient harms. It supports a reorientation away from formal principles and towards human welfare and a person’s capacity to flourish.

\subsection{As a Way of Understanding Responsibilities and Duties}

The moral responsibilities entailed by human rights treaties apply to a wide range of actors and environments. States are often the primary duty-bearers, with respect for human rights being a key element of regime legitimacy \cite{rawls1993law}.  Individuals also have duties not to violate human rights and there are a number of national and supranational mechanisms, such as the ICC and the European Court of Human Rights, to address this. Increasingly, human rights are also understood to apply to companies or organizations and to those employed by them --- including for people working in the technology sector. These frameworks can therefore help us to understand who has a moral responsibility to do what, serving as a well-spring for responsible innovation in AI.

To begin with, human rights have ramifications for the way in which scientific and commercial research is conducted, especially when research concerns human subjects. This connection between bioethics and human rights was explicitly recognized in the aftermath of the Holocaust \cite{baker2001bioethics}, leading to the development of safeguards for participants such as those described in the 1964 \textit{Helsinki Declaration} and the 1979  \textit{the Belmont Report}.\footnote{\url{https://www.hhs.gov/ohrp/regulations-and-policy/belmont-report/read-the-belmont-report}} More recently, the \textit{the Universal Declaration on Bioethics and Human Rights (UDBHR)},\footnote{\url{https://en.unesco.org/themes/ethics-science-and-technology/bioethics-and-human-rights}}
adopted by UNESCO's General Conference in 2005, further affirmed the importance of human dignity, human rights and fundamental freedoms when it comes to research. Given recent controversy surrounding numerous publications in the field of ML and existing data collection practices, these principles, and the institutional review protocols they necessitate, serve as a valuable precedent for the broader ML research community \cite{van2020ethical}.

The focus on human rights can also help us understand the responsibilities technology companies owe within the wider ecosystem of duty bearers. The 2011 United Nations Guiding Principles on Business and Human Rights (UNGP) outline the specific responsibility of businesses to respect human rights, including by identifying, preventing, and mitigating salient human rights risks \cite{raso2018artificial}. As the UNGPs make clear, when it comes to human rights, it is important to move beyond good intentions: those developing new technologies need to make an informed effort to understand the implications that a technology will have for rights holders --- and to put in place measures that ensure that the rights are upheld through processes of evaluation, review and assessment.



\subsection{As a Framework for Civil Society Participation}

A major challenge for the AI ethics community centers upon the inclusion of relevant voices when addressing the design and governance of AI systems, and the cultivation of an effective \textit{lingua franca} --- or participatory processes --- that make it easier for needs of historically marginalized communities to be articulated and for their claims to be effectively met. Against this backdrop, a key cluster of human rights centers upon and recognize the value of wide participation. For example, Articles 20 and 21 of the UDHR guarantee ``freedom of peaceful assembly and association'' and ``freedom to participate in political processes'' respectively. Moreover, the idea that people are ``rights-holders'' serves as a basis for engagement and source of authority that is quite different from engagement as ``consumers'', ``citizen[s]'', ``stakeholders'' or ``affected parties''. Taken together, these two interrelated elements of participation and empowerment explain why approaches towards responsible AI, pioneered by civil society organizations, have tended to invoke human rights more often than fairness research situated within the machine learning community. 

For instance, the \textit{Mapping Consensus in Ethical and Rights-based Approaches to Principles for AI} report published by the Berkman Klein Center for Internet \& Society \cite{fjeld2020principled} found that among thirty-six different sets of AI principles published by private and public agencies, human rights are a major focus --- with civil society and trans-national governmental agencies relying most heavily on this framework. Documents drafted by trans-national governmental agencies such as \textit{AI for Europe} by the European Commission and \textit{Ethics Guidelines for Trustworthy AI} by the European High Level Expert Group on AI also foreground human rights. And out of the five civil society-drafted AI principles documents, three of them --- \textit{Toronto Declaration} by Amnesty International \& Access Now,\footnote{\url{https://www.accessnow.org/the-toronto-declaration-protecting-the-rights-to-equality-and-non-discrimination-in-machine-learning-systems}} \textit{Universal Guidelines for AI} by The Public Voice Coalition,\footnote{\url{https://thepublicvoice.org/ai-universal-guidelines/}} and \textit{Human Rights in the Age of AI} by Access Now,\footnote{\url{https://www.accessnow.org/cms/assets/uploads/2018/11/AI-and-Human-Rights.pdf}} --- explicitly adopt a human rights framework, while a fourth one, \textit{Top 10 Principles for Ethical AI} by UNI Global Union,\footnote{\url{http://www.thefutureworldofwork.org/opinions/10-principles-for-ethical-ai/}} includes discussion of human rights risks. 

By way of contrast, research done within the ML fairness, accountability, transparency and ethics research community rarely invokes a rights-based framework. Drawing upon a survey of 138 papers/abstracts published in the FAT* conference in 2019 and 2020, we found only one research paper \cite{kaminski2020multi} and 2 tutorial abstracts \cite{cath2020leap,szymielewicz2020where} that engage with the human rights scholarship.\footnote{12 other research papers mention the phrase \textit{human rights}, but do not engage with it beyond that.} More precisely \cite{kaminski2020multi} proposes an impact assessment methodology that they situate within the human rights assessment literature, \cite{cath2020leap} conducted a translation tutorial between human rights scholarship and FATE research and \cite{szymielewicz2020where} conducted a hands-on tutorial where they used a human rights framing to test academic concepts and their formulation in policy initiatives around algorithmic accountability and explainability.

In addition to the substantive value of the human rights framework, the salience that human rights have both for people affected by new technologies and in policy circles provides additional reason to close this significant gap in the AI FATE literature. For AI researchers to build effective partnerships with civil society, bridging work needs to be done. 
In certain domains bridging work is already underway, for instance between the community of NLP researchers working in the space of detecting online abuse and the RightsCon community \cite{prabhakaran2020online}. Yet, failure to engage with human rights scholarship more widely, risks leading to a situation in which FATE researchers end up ‘speaking a different language’ from those their products affect, making it harder to conduct participatory research with civil society organizations and foregoing a major opportunity to strengthen the practice of AI ethics more widely.

\section{Illustrations}
\label{sec_illustrations}

The potential impact of AI on human rights is wide-ranging, something that is recognized by recent commentary on the human right to science by the UN Economic and Social Council which notes  ‘applications of artificial intelligence in industry or services can lead to enormous gains in productivity and efficiency’ while also expressing concern that algorithms could be incorporated into weapon systems or used to reinforce discrimination.\footnote{\url{https://undocs.org/E/C.12/GC/25}} Rather than attempt to catalogue the full range of impacts AI might have, we focus in this section on three human rights in particular, to show what a human rights framework may add to the responsible AI discussion.  For each right, we walk through specific examples of how the values at play may influence decisions in an algorithmic context.

\subsection{The Human Right to Freedom from Discrimination}


The right to be free from discrimination is a negative right not to be harmed in certain ways, and it is heavily impacted by prevalent forms of algorithmic bias.
This right is enshrined in various universal and regional legal instruments of human rights, and forms one of the core rights in the UDHR. In particular, article 2 of the UDHR states: “everyone is entitled to all the rights and freedoms set forth in this Declaration without distinction of any kind, such as race, colour, sex, language, religion, political or other opinion, national or social origin, property, birth or other status”, essentially extending all the rights enshrined in the declaration to all humans without discrimination. 

The right to be free from discrimination is also often the right that is most directly relevant to a majority of work in the FATE community. Fairness research has identified numerous types of biases in various algorithmic systems \cite{barocas2017fairness,blodgett2020language}. An algorithmic system that treats individuals differently based on an attribute such as race or gender, with negative consequences for group members and without due cause, is in and of itself a violation of the right to be free of discrimination. However, certain instances of discrimination that result in withholding other rights have a compounding effect. For instance, an algorithmic content moderation system that disproportionately censors individuals speaking a certain dialect \cite{sap2019risk} not only risks interfering with their right to freedom of expression, but also potentially impedes their right to be free from discrimination in exercising that right. 

Since early research into algorithmic fairness dealt with the applications of AI in US law enforcement such as predictive policing and recidivism prediction, and regulations around anti-discrimination in housing, loans, and education in the US, the FATE community’s inquiries into this space draws largely on US legal frameworks such as the Civil Rights Acts and Fair Housing Act, as well as on US legal concepts of discrimination \cite{green2020algorithmic}. One side effect of this is that the scope of this conversation has also been largely limited to discrimination in the US context, compared to the more global human rights framework. For instance, while the right to be free of discrimination based on religion or language has equal standing within the global human rights framework, these axes of discrimination are rarely dealt with within the FATE research community, compared to discrimination based on race or gender, which are prominent concerns in the US American public discourse. This gap, in terms of understanding the full range of characteristics, that may serve as axes of unjust discrimination, can be addressed, in part, through reflection on the more expansive categorisation invoked by the UDHR.

Furthermore, indexing FATE research on the legal framework of a particular country carries with it additional risk, when viewed form a global standpoint. For example, a single country’s legal frameworks may not give all groups adequate protection against discrimination, an issue that looms particularly large in the context of unjust laws or national practices. To guard against these pitfalls, a more universal set of principles such as the universal human rights doctrine, may serve as a useful reference point, and as means of checking national laws this kind of gaps or denial of equal rights to citizens.

\subsection{The Human Right to Health}

The human right to health enshrines “the right of everyone to the enjoyment of the highest attainable standard of physical and mental health” (ICESAR, 1966, Art. 12.1).\footnote{\url{https://www.ohchr.org/en/professionalinterest/pages/cescr.aspx}}  It works primarily as a positive right that creates a duty for states to lower infant mortality, promote child development, provide medical services to their populations, and share medical knowledge, among other things. Given the growth of ML-enabled services and diagnostic tools within the healthcare sector \cite{de2016automated,yu2018artificial},
AI research has the potential to intersect with this right in important ways. 

The right to health grounds an entitlement to access health care on terms that are free from discrimination, with particular attention being paid to protected groups such as women and those with physical disabilities. Given evidence that racial bias affects algorithms deployed in a healthcare context \cite{obermeyer2019dissecting}, measures to mitigate the harms that biases can create prior to an algorithm’s deployment can be understood as a human rights obligation. Additionally, the duties that correspond to the right to health make reference to that   “highest attainable standard” of care, a specification that acknowledges that the content of the right will vary dynamically according to time and place. For example, governments have an obligation to give their citizens the highest attainable standard of health care. But if the country is very poor, then the required standard of treatment might still be quite low. Contrastingly, if AI services can be used to bring down the cost of healthcare, or the lack of access to it, then what is 'attainable' for the country may begin to rise, and the right to health may become a right to access and benefit from AI services. This invites us to think about the relationship between healthcare and AI in a different light --- not only as a source of potential harms but through the lens of human rights-enabling technology. 

Current estimates suggest that there are not enough trained doctors and physicians globally to provide everyone in the world with a high level of medical care, a problem that is particularly true in low-income countries \cite{scheffler2008forecasting}. This shortfall could be addressed through the global redistribution of economic resources \cite{caney2006global}. However, it is also something that AI researchers are in a special position to influence, through the creation of diagnostic tools that can be deployed affordably at scale and the development of customized digital healthcare services. For example, ML-enabled technology is now being used to detect diabetic retinopathy at scale in India \cite{yu2018artificial}. A similar point, about the creation of potentially low-cost tools and services, holds true for the human right to education. AI-enabled services could, in principle, make it possible for many more children to enjoy customised education in their own language, thereby improving global literacy and learning.

In both cases it is important to steer clear of the pitfalls of technological solutionism \cite{morozov2013save}. As human rights advocates note, the exercise of human rights and impediments to them are frequently political in nature \cite{sen2005human,gabriel2019effective}. Yet these opportunities also ground a positive aspiration for AI: that it will expand the feasibility frontier so that people can enjoy a higher standard of human rights fulfilment around the world. Moreover, this goal dovetails with the  Sustainable Development Goals \cite{tomavsev2020ai} and would likely find widespread support among those who experience limited access to services.

\subsection{The Human Right to Share in Scientific Advancement}

The human right to science states that everyone has a right to “share in scientific advancement and its benefits” (UDHR, Art. 27). Science is understood here to include: (1) knowledge, (2) the application of that knowledge, and (3) the method of the knowledge production. Moreover, the right applies both to scientific knowledge itself and to the benefits it creates. This right has special relevance for AI both because it can be understood as a scientific practice and also because of the way in which AI is increasingly used to advance scientific progress, as with DeepMind’s AlphaFold which successfully predicted the structure of almost every known protein \cite{senior2020improved}.

The human right to science advances an ideal of inclusive science and prohibits discrimination both among those employed in scientific pursuits and among its beneficiaries. On this point Michelle Bachelet, UN High Commissioner for Human Rights affirms that “those participating in the global scientific effort should… take into account the needs and experiences of women, members of minority communities, Indigenous scholars, persons with disabilities, people living in poverty and people living in less developed countries – among others. Only then will research fully address all communities – and contribute to reducing the unequal access to scientific developments and capabilities across different countries and regions”.\footnote{\url{https://www.ohchr.org/EN/HRBodies/HRC/Pages/NewsDetail.aspx?NewsID=26433&LangID=E}} This element of the right to science underscores the need to systematically promote diversity, equity and inclusion in AI research.

Significantly, the human right to science also bears upon the distribution of benefits enabled by science. The UDHR clearly states that the right is to be interpreted in a way that respects the intellectual property of researchers. However, it situates these “moral and material interests” within the wider aspiration that science should be geared towards fulfilment of human rights. Speaking on behalf of CERN and the WHO among others, Bachelet states that “the benefits of scientific and medical progress were always meant to be shared. The great beauty of science is that it has no borders – and that, working together, every scientist and student of science can contribute to the shared knowledge and benefit of all”. In the context of AI research, the human right to science asserts the importance of this technology ultimately benefiting a large section of humanity, including those historically excluded from the benefits of scientific advances.  Without the assersion of this right, scientific progress may otherwise continue to exclude insights, ideas, and concerns from people who are historically excluded, pulling AI research farther from a trajectory that supports the global social good.

\ifcasestudyinmaintext
\section{Worked Example: Online Content Moderation}
\label{sec_example}


Having outlined how some of the specific rights are of core relevance to AI ethics conversations, we now dive more deeply into a specific problem domain to demonstrate how the consideration of human rights in AI based interventions can be applied in practice. More specifically, we use the application domain of online content moderation, focusing on a hypothetical platform where people write and read one another's text. The domain of content moderation is an incredibly complex one, with challenges around detecting online abuse and toxic content at scale and across geo-cultural contexts, while also accounting for conversational contexts, and a multitude of intervention approaches towards content removal or demotion, as well as concerns on transparency and accountability \cite{prabhakaran2020online}. Our choice of this domain is motivated by these complexities, as it helps to demonstrate the various competing considerations involved, including different sets of rights-holders, and the different kinds of risks to their rights. 
This section is not meant as a blueprint of a solution to tackle this complex problem, rather as a way to show how a human rights based approach illuminates some of these competing considerations, and centers the conversation around these considerations that are often overlooked in traditional approaches. Also, note that this discussion is not limited to just the three rights discussed as illustrative examples above, rather we draw from the full set of UDHR rights.  



\subsection{Rights-holders}

One important way a human rights-centered perspective changes the approaches towards responsible AI is that it forces us to confront the question of whose rights are at risk and what those risks are. As Blodgett \cite{blodgett2020language} points out, research on fairness failures in AI tend to sidestep answering these questions. In the context of our current work, this pushes us to distinguish \textit{rights-holders}, i.e., those whose rights are at risk, from stakeholders, i.e., anyone who can claim a stake in the system.  In the case of online content moderation, the rights-holders include:
\begin{itemize}
    \item the \creators of the content,
    \item the \addressees to whom the content is specifically directed,
    \item the \audience on the platform who are exposed to the content, and
    \item the \referrents who are implied or denoted in the content,
    \item the \moderators who may have to review the content.
\end{itemize}

\noindent Moreover, the rights-holders can be an individual, a set of individuals, a population subgroup, or a community of people. The platform owners are a stakeholder, but they are not relevant rights-holders in this example since their human rights are not at risk in these scenarios which focus on online communication.



In the rest of this section, we incrementally analyze different scenarios, increasing the complexity of an example text content moderation system that is used on a platform for online communication.  We consider the relevant rights and rights-holders given the \textit{presence} of problematic online text (Scenario 1), followed by the relevant rights when \textit{automatic content moderation} is applied (Scenario 2), then how individuals' rights are affected when addressing \textit{biases in text content moderation models} (Scenario 3), and finally with the addition of \textit{human-in-the-loop content moderation} (Scenario 4). We primarily focus on the rights and right-holders with respect to the set of human rights enshrined in the UDHR, however particular contexts might require considerations of additional sets of rights, e.g., the UN Declaration on the Rights of Indigenous Peoples may be more appropriate in the context of an AI intervention where indigenous people are a rights-holder. 

A summary of some of the relevant UDHR rights at play with respect to each right-holder in this example is presented in Figure \ref{tab:case_study_rights}.  \addressees, \audience, \creators, and \referrents are affected differently depending on how the rights are prioritized.  Interventions to tackle harmful content online must attempt to untangle these tensions and mitigate the risks, balancing the rights at play.


\begin{table}
    \centering
    \small
\begin{tabular}{|l|p{.42\textwidth}|llll|}
\hline
    \textbf{Right} & \textbf{Abridged Description} & \multicolumn{4}{|c|}{\textbf{Rights-holders}} \\\hline\hline
UDHR\S5&\textit{No one shall be subjected to torture or to cruel, inhuman or degrading treatment or punishment. }&    {\sc Addressees}  &  &   &  \\\hline
 UDHR\S1 & \textit{People should act towards one another in a spirit of brotherhood} &{\sc Addressees} & &   {\sc Creators}& \\\hline
 UDHR\S3 & \textit{Everyone has the right to...liberty and the security of person.} & {\sc Addressees} &  &  {\sc Creators} &  {\sc Referrents} \\\hline
 UDHR\S25 & \textit{Everyone has the right to a standard of living adequate for the health and well-being of himself and of his family.}&  {\sc Addressees} &  & & {\sc Referrents} \\
UDHR\S12 & \textit{No one shall be subjected to arbitrary interference with his privacy, family, home or correspondence, nor to attacks upon his honour and reputation.} & &&&\\\hline
 UDHR\S20 & \textit{Everyone has the right to freedom of peaceful assembly and association.}&  {\sc Addressees} & {\sc Audience}  &  &   {\sc Referrents}    \\\hline
UDHR\S7 & \textit{...All are entitled to equal protection against any discrimination...[and] incitement to such discrimination.} &{\sc Addressees}  & {\sc Audience} &   &  \\\hline
 UDHR\S18 & \textit{Everyone has the right to freedom of thought...[and to] manifest his religion or belief in teaching, practice...} &{\sc Addressees}  & {\sc Audience}& {\sc creators}  &       \\
  UDHR\S19 & \textit{Everyone has the right to freedom of opinion and expression...includes freedom to hold opinions without interference and to seek, receive and impart information and ideas through any media...}&   & &   &   \\\hline
   UDHR\S29 & \textit{Everyone has duties to the community in which alone the free and full development of his personality is possible.}&{\sc Addressees} &{\sc Audience}&{\sc Creators} &   {\sc Referrents} \\\hline
 UDHR\S27 & \textit{Everyone has the right freely to participate in the cultural life of the community...to the protection of the moral and material interests resulting from any...[literary production] of which he is the author. }&&&{\sc Creators} &   \\\hline
UDHR\S23 & \textit{Everyone has the right to...just and favourable conditions of work...to form and to join trade unions for the protection of his interests.} &  \multicolumn{4}{|c|}{\sc Moderators} \\\hline
 \end{tabular}
    \caption{Abridged rights relevant to the different parties in an online communication platform with content moderation (non-exhaustive).}
    \label{tab:case_study_rights}
 \end{table}


\noindentparagraph{\bf Scenario 1: Presence of Harmful Content}

The \creators who produce harmful content are owners of the content, describing the \referrents and communicating to the \addressees in front of an \audience. In producing content, \creators are exercising their rights for freedom of opinion and expression (UHDR\S18,19).  Yet, determining which content may be considered ``harmful'' can stem from whether it is at odds with the need to respect 
the human rights of others, i.e., 
whether the content causes degrading treatment (UDHR\S5).

If a \creator's content reveals private and sensitive information about a specific individual, that individual is a \referrent whose rights to privacy and reputation (UDHR\S12) is in question. If a \creator's content spreads dangerous misinformation about a community, the \referrents are everyone in the community, and may violate their right to security (UDHR\S3). \addressees have rights similar to the \referrents, which additionally include rights for security (UDHR\S3) against anyone who might harm them (such as an incited \audience) and a standard of living that is adequate for well-being (UDHR\S25). 
If it may be said that people `assemble' on an online platform, harmful content can additionally touch on the freedom to peaceful assembly and association (UDHR\S20) for \addressees, \audience and \referrents, as well as the \creators, albeit in different ways. 

\noindentparagraph{\bf Scenario 2: Automatic content moderation}
Now, let us consider a scenario where a text-based content moderation system is introduced on the platform to remove content that is assessed as toxic or offensive. The task of removing harmful content may be at odds with the rights of \creators to freely participate in the culture of a community (UDHR\S27), to freedom of thought (UDHR\S18), and to freely impart opinions through any media (UDHR\S19), as well as the corresponding rights of the \audience to receive information and ideas without interference (UDHR\S19). 

Moreover, like any machine learning-based intervention, text-based censoring interventions make errors.  
When such a model makes a false positive prediction, i.e., incorrectly labeling harmless content to be toxic, it risks interfering with the \creators' right to freedom of expression without undue interference (UDHR\S19). A false negative prediction, where harmful content is shared, also has the potential to invoke all the rights described in Scenario 1. 
It is important to note, however, that the automatic content moderation does mitigate this risk in a majority of cases where content was removed through a true positive prediction by the model.

\noindentparagraph{\bf Scenario 3: Biases in NLP models}

An additional factor in automatic content moderation is the existence of biases in the NLP models that are used to detect offensive content \cite{dixon2018measuring,sap2019risk,hutchinson2020social}.  The presence of certain lexical items associated with certain people or groups of people can cause the model to predict, for example, a higher toxicity score, resulting in removal of content that should not be removed. Such biases have been documented around mentions of LGBTQ+ identity terms \cite{dixon2018measuring}, mentions of people with disabilities \cite{hutchinson2020social}, mentions of controversial people \cite{prabhakaran-etal-2019-perturbation}, as well as dialectal speech \cite{sap2019risk}. For instance, the sentence \textit{I am a deaf person} is assigned a toxicity score of 0.44 while \textit{I am a person} is assigned a score of 0.08 by a commonly used automatic content moderator tool Perspective API.\footnote{\url{https://www.perspectiveapi.com/}} The same tool assigns a toxicity score of 0.90 for the sentence \textit{I hate Justin Timberlake}, compared to a score of 0.69 for \textit{I hate Rihanna}. The impact of such biases is that messages with certain features, often associated with certain individuals or communities are disproportionately censored. While detecting and mitigating such biases is an active research area within the NLP community lately, much of this work does not engage with why certain biases are harmful, or who it harms \cite{blodgett2020language}. A human rights based approach to studying such biases centers the conversation around whose rights are at risk and how. 

In the cases where biases incorrectly flag messages with references to groups or individuals (e.g., using names or identity terms) as harmful, what is often discussed are the harms it causes to those individuals or groups. 
In particular, such biases would mean that fewer mentions of these terms will pass the automated content moderation, lowering the inclusion of these \referrents in the online content, posing risks to their right to be free from discrimination (UDHR\S2, UDHR\S7).
This is especially problematic in certain cases, for instance, mentions of disabilities, exacerbating the already reduced visibility of disability in the public discourse, further reducing the public awareness of its prevalence and negatively influencing societal attitudes towards these people \cite{scior2011public}. These effects can then have knock-on effects on the ability of minorities to exercise their rights, for example, due process and access to justice. So the issue is not only about whether human rights are directly infringed but also about whether technology is creating an environment in which it is easier or harder for rights to be successfully exercised.

As discussed in Scenario 2, these biases also pose risks to \creators' right to freedom of expression (UDHR\S19), however the group of people whose rights are at risk differs depending on the kind of bias. For instance, since people with disabilities are also more likely to talk about disability, increased censorship on content about ability status could arguably limit the right of people with different disabilities to participate in public fora and seek opinions on this topic (UDHR\S19). However, this is not always true: a bias around a person's name may not pose risk to their freedom of expression, rather of those who are more likely to write about them. In most cases, such a group of people may not belong to a certain protected group, and hence there may not be a risk to the right to be free of discrimination (UDHR\S2), but depending on the particular individual, for instance, biases around a certain religious leader might result in disproportionate removal of content of that religious group. In contrast, the biases around certain dialects or other linguistic features associated with certain protected groups poses a direct risk to the right to be free of discrimination (UDHR\S2), especially the right to freedom of expression (UDHR\S19). For instance, the biases documented by \cite{sap2019risk} that African American Vernacular English (AAVE) is more likely to be labeled as toxic will result in disproportionately censoring \creators belonging to that community.

\noindentparagraph{\bf Scenario 4: Human-in-the-loop Content Moderation}
Now, let us consider a human-in-the-loop scenario where an NLP-based content moderation system first flags the content to be reviewed, which is then routed to human content \moderators.  As discussed in the last scenario, the rights of \creators, \addressees, \referrents, and \audience can be affected by errors in an automatic content moderation system. When \moderators are looped into this pipeline examine the correctness of flagged content and make final decisions on what should be excluded or included, an argument can be made that a model's bias towards false positives for some \creators, \addressees, or \referrents could actually result in enhanced and expedited safety within such a human-in-the-loops scenario, as there would be a higher and faster scrutiny of problematic messages concerning them (in a scenario where content with higher score is routed to the moderator first). 

On the other hand, humans are not error-free, have their own biases, and are affected by automation bias \cite{goddard2012automation},
potentially making them more likely to agree with a biased system's errors. This may affect different subgroups differently and widen discrepancies in the rights of different actors. Similarly, since model scores may sometimes be used to select and prioritize messages for review by moderators \cite{veglis2014moderation,jurgens2019just}, the decisions on whether to review models with higher or lower scores first will determine which rights-holders are impacted, and which rights are at risk.
Human-in-the-loop settings also mandates consideration of the human rights of the \moderators, whose right to have safety at work (UDHR\S23) may be at risk, due to the problem of continuous exposure to distressing and toxic content \cite{roberts2016commercial}. Furthermore, concerns around crowdwork practices, including fair remuneration for the \moderators will also have to be taken into account \cite{gray2019ghost}, as ``just and favourable conditions of work'' are also enshrined in (UDHR\S23).

\fi

\section{Discussion}
\label{sec_recco}

In this paper, we have sought to show how greater attention to human rights can help ground the aspiration to build more ethical AI systems, anchored in values that have a measure of cross-cultural affirmation. Orienting the research in this space away from the machines and the risks of their biases, and towards humans and the risks to their rights, can help center conversation around the harms caused by a technology, including specific consideration of who is harmed, and how those harms may be mitigated. This reframing also has the potential to better align efforts by the FATE community to improve AI systems with the wider global advocacy movement that is committed to securing human rights and their fulfilment.


In support of these goals, future research on human rights-based approaches towards to AI ethics could help bridge the gap in multiple ways. First, there is a need for \textbf{translational} research in this space that can address, more precisely, how human rights principles map to the current ethics-based considerations in AI, such as fairness, consent, privacy, and ownership. Such research could be aimed at building a shared vocabulary of concerns, values, and expected outcomes as an important first step for meaningful bridging between computer science researchers and civil society activists working in this space. This should crucially include clarifying the needs of civil society that are overlooked and potentially easily addressed through technology, as well as the challenges of ensuring fairness of algorithmic predictions at scale.  

Another line of work could look into the \textbf{functional} aspects of a human rights based inquiry into AI ethics. For instance, what does a human rights based inquiry into ML fairness reveal that existing methodologies do not. 
\ifcasestudyinmaintext
The worked example presented in Section~\ref{sec_example} in the context of online content moderation demonstrates some of these functional aspects of a human rights based approach. In particular, it pushes us to identify the rights holders, to identify which of their rights are at risk, and to map out how those risks interact with the claims of other right holders --- making it essential for AI researchers to contend with various trade-offs, when determining how to intervene. For instance, fairness researchers need to consider how bias mitigation measures, designed to mitigate risks to the human rights of \creators, might create or increase risks to certain other rights of the \audience. 
\else
The worked example in Appendix~\ref{sec_example} in the context of online content moderation demonstrates some of these functional aspects of a human rights based approach. In particular, this approach pushes us to identify the rights holders, to identify which of their rights are at risk, and to map out how those risks interact with the claims of other right holders. Thus, it is essential for AI researchers to contend with various trade-offs, when determining how to intervene. For instance, fairness researchers need to consider how bias mitigation measures, designed to mitigate risks to the human rights of content creators, might create or increase risks to certain other rights of the audience. 
\fi


Finally, advances in AI have the potential to play an important role in \textbf{enabling} stronger human rights fulfilment around the world. As we lay out in Section~\ref{sec_illustrations}, AI has already been shown promise when it comes to extending the scope and content of various human rights (such as the right to health and right to scientific advancements) to marginalized communities. Future research should consider what role AI can play in enhancing access to health and education for communities in lower-income countries. Similarly, AI-based technologies such as automatic captioning might play an important role in increasing access to education for people with disabilities.
There is more work to be done in this space, not only in employing AI in rights-enabling applications, but also in bringing the advancements in AI and technology to communities around the world, rather than keeping it only within the reach of a select few.

\section{Conclusion}



AI ethics does not always give due recognition to the idea that every human life has value and also that human life is fragile -- considerations that ground a set of important moral claims on institutions, new technologies, and on one another. The notion that people have human rights builds upon this foundation, recognizing that we all share certain vulnerabilities, that we ought not to be harmed, and these considerations guide how AI should be developed. In this paper we have suggested that human rights-based considerations can perform three valuable functions in the context of AI research. First, human rights can serve as a partial basis for AI value alignment across a range of cultures and different contexts, due to the qualified but significant cross-cultural validity that they evidence. Second, a human rights framework can help us understand how ethical principles governing the design and deployment of AI systems translate into different responsibilities for the actors that comprise different parts of the AI ecosystem. Third, human rights can function as a language that enables deeper collaboration between AI researchers, civil society groups and the people impacted by these technologies. In this way it can help close the gap between technical research focusing on algorithmic fairness, and the claims of those who interact with AI systems on the ground and in the public sphere. Taken together, these elements of human rights doctrine make it an appealing set of guiding principles for AI researchers and practitioners to draw upon.


\section*{Acknowledgements}

We thank Roya Pakzad, Jamila Smith-Loud, Tan Zhi Xuan, and Ben Zevenbergen for helpful conversations on this topic and for useful feedback on early drafts of this paper. We also thank the
anonymous reviewers for their constructive feedback.

\bibliographystyle{ACM-Reference-Format}
\bibliography{eaamo_main}

\ifcasestudyinmaintext
\else
\section*{APPENDIX}
\appendix

\fi

\end{document}